%% file: draft.tex
\documentclass[11pt]{article}

\usepackage[final]{acl}
\usepackage{times}
\usepackage{latexsym}

\usepackage[T1]{fontenc}

\usepackage[utf8]{inputenc}

\usepackage{microtype}

\input{preamble}
\usepackage{subfigure}
\usepackage{nicefrac}

\DeclareMathAlphabet{\mathbcal}{OMS}{cmsy}{b}{n}
\usepackage{amsmath}
\usepackage{mathrsfs}
\usepackage{comment}

\usepackage{bm}
\usepackage{bbm}
\usepackage{amssymb}
\usepackage{enumitem}

\usepackage{xcolor}
\usepackage{wrapfig,lipsum,booktabs}

\usepackage[strict]{changepage}
\usepackage{float}
\usepackage{framed}

\usepackage{inconsolata}

\usepackage[strict]{changepage}

\usepackage{framed}

\usepackage{natbib}

\usepackage{graphicx}
\graphicspath{{final/Images/}}

\definecolor{formalshade}{rgb}{0.95,0.95,1}

\usepackage{tabularx}
\usepackage{colortbl}
\usepackage{subfiles}
\usepackage{comment}

\newcommand{\ryan}[1]{\textcolor{blue}{~Ryan:~#1}}

\newcommand{\lead}[1]{\textcolor{purple}{Lead:~#1}}

\renewcommand{\ryan}[1]{}
\renewcommand{\lead}[1]{}

\title{Forecasting Time Series with LLMs via \\Patch-Based Prompting and Decomposition}

\author{
\textbf{Mayank Bumb}$^{1}$, 
\textbf{Anshul Vemulapalli}$^{1}$, 
\textbf{Sri Harsha Jella}$^{1}$, 
\textbf{Anish Gupta}$^{1}$, \\
\textbf{An La}$^{1}$, 
\textbf{Ryan Rossi}$^{2}$, 
\textbf{Hongjie Chen}$^{3}$, 
\textbf{Franck Dernoncourt}$^{2}$,
\textbf{Nesreen Ahmed}$^{4}$, 
\textbf{Yu Wang}$^{5}$ \\
$^{1}$University of Massachusetts Amherst, 
$^{2}$Adobe, 
$^{3}$Dolby Labs,
$^{4}$Intel,
$^{5}$University of Oregon\\
}

\begin{document}
\maketitle

\begin{abstract}
Recent advances in Large Language Models (LLMs) have demonstrated new possibilities for accurate and efficient time series analysis, but prior work often required heavy fine-tuning and/or ignored inter-series correlations. In this work, we explore simple and flexible prompt-based strategies that enable LLMs to perform time series forecasting without extensive retraining or the use of a complex external architecture.
Through the exploration of specialized prompting methods that leverage time series decomposition, patch-based tokenization, and similarity-based neighbor augmentation, we find that it is possible to enhance LLM forecasting quality while maintaining simplicity and requiring minimal preprocessing of data. To this end, we propose our own method, PatchInstruct, which enables LLMs to make precise and effective predictions. 
\end{abstract}

\input{draft/1.intro}

\input{draft/2.related_work}

\input{draft/3.method}

\input{draft/4.1.dataset}

\input{draft/4.2.exp}

\bibliography{draft}
\appendix
\input{draft/appendix}

\bibliographystyle{acl_natbib}
\end{document}

%% file: preamble.tex
\usepackage{xcolor}
\definecolor{thedarkblue}{RGB}{0,0,120} %
\definecolor{mydarkblue}{rgb}{0,0.08,0.45} %
\definecolor{darkblue}{rgb}{0,0.08,180}
\colorlet{TufteRed}{red!80!black}
\definecolor{theblue}{RGB}{0,0,180}
\colorlet{thered}{TufteRed}
      
\usepackage{microtype}
\usepackage{balance}

\usepackage{booktabs}
\usepackage{tabularx}

\usepackage{amsmath,amssymb,amsthm}

\newcommand{\eat}[1]{\ignorespaces}
\usepackage{comment}

\newcommand{\journal}[1]{} %

\usepackage{tikz}
\usepackage{verbatim}
\usetikzlibrary{arrows}
\usetikzlibrary{shapes,snakes}
\usetikzlibrary{decorations.pathmorphing} %
\usetikzlibrary{fit}					%
\usetikzlibrary{backgrounds}	%

\usepackage{ragged2e}
\usepackage{multirow}
\usepackage{microtype}
\usepackage{balance}
\usepackage{setspace}

\graphicspath{{./}{./graphics/}}
\newcolumntype{H}{>{\setbox0=\hbox\bgroup}c<{\egroup}@{}}

\newcolumntype{R}[1]{>{\RaggedLeft\arraybackslash}} %
\newcolumntype{L}[1]{>{\RaggedRight\arraybackslash}} %

\AtBeginEnvironment{pmatrix}{\setlength{\arraycolsep}{2pt}}

\DeclareMathOperator{\hugeE}{\mbox{\huge\raise-0.3ex\hbox{E}}}
\DeclareMathOperator{\p}{\mathbb{P}}
\DeclareMathOperator{\hugep}{\mbox{\huge\raise-0.3ex\hbox{$\p$}}}

\usepackage[most]{tcolorbox}
\usepackage{amsmath}  %

\newtcolorbox{PromptTextBox}[1]{%
  colback=gray!5,
  colframe=black,
  boxrule=0.5pt,
  arc=2mm,
  width=\linewidth,
  title={#1},
  fonttitle=\bfseries,
  enhanced,
  left=2mm,
  right=2mm,
  top=1mm,
  bottom=1mm,
  boxsep=1mm
}

%% file: draft/1.intro.tex
\section{Introduction}

Time-series forecasting (TSF) has a broad range of applications in agriculture, business, epidemiology, finance, etc. Many of these applications require robust predictions of time series, and accurately modeling the dependencies between variables remains to be a challenge \cite{shao2020time}. Traditional forecasting models such as ARIMA, LSTMs, and even Transformer/Graph-based architectures have displayed a strong performance on these tasks \cite{zhou2024ditto}.

More recently, Large Language Models (LLMs) have shown a promising future in modeling time series, with accurate predictions that rival state of the art (SOTA) methods, due to their strengths in pattern recognition, sequence modeling, and generalization across tasks. However, current LLM-based methods often rely on complex architectures or require heavy fine-tuning, limiting their scalability to real-world applications.

One prominent approach, S$^2$IP-LLM~\cite{pan2024textbfs2ipllmsemanticspaceinformed}, embeds time series into a semantic space to enhance forecasting performance. While effective, it introduces two key limitations. First, it incurs a high computational cost during inference due to its reliance on complex decomposition and patching pipelines. Second, it does not explicitly model dependencies across related time series, which can be critical in domains such as traffic and energy forecasting where inter-series relationships play a significant role.

\begin{figure*}[t]
  \centering
  \includegraphics[scale=0.65]{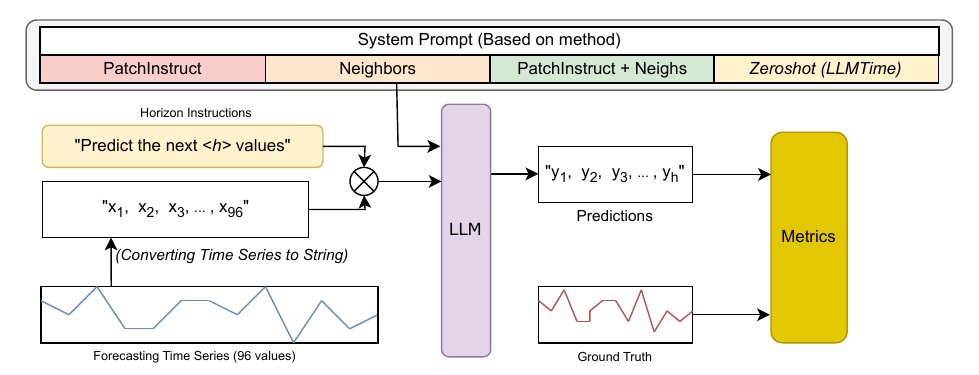}
  \caption{LLM-based Time-Series Forecasting Pipeline}
  \label{fig:overall_figure}
\end{figure*}

We aim to develop a method (see Figure \ref{fig:overall_figure}) that maintains the predictive strength of LLM-based models while addressing the above limitations of inference speed and generalization. Therefore we guide our experimentation around the idea of whether we can create general-purpose prompts that guide LLMs to forecast time series both accurately and efficiently, without requiring model fine-tuning or architectural changes.

To this end, we introduce PatchInstruct, a prompt-based framework that tokenizes time series data into meaningful patches that encapsulate temporally relevant patterns and guides the LLM via structured natural language instructions to output precise predictions. Unlike prior work, PatchInstruct requires no model retraining or architecture modification and also significantly reduces inference time (in comparison to the baseline and complex architectures) alongside token usage while preserving or improving accuracy.

We compare PatchInstruct with several other prompting strategies—including Zero-shot, Neighbors, and PatchInstruct + Neighbors—and evaluate them on diverse, real-world datasets (Weather and Traffic), primarily using GPT-4 and GPT-4o as the LLM backbones.

PatchInstruct based methods consistently outperform baselines across small forecasting horizons and datasets. Notably, PatchInstruct almost always achieves top forecasting accuracy on small values of horizons (at most $12$), in regards to MSE/MAE, while reducing the inference overhead by 10x–100x compared to S$^2$IP-LLM while maintaining strong accuracy.

These results support our hypothesis that prompt engineering can effectively replace some degree of architectural complexity and will help with scalable and domain-adaptable time series forecasting using LLMs.

%% file: draft/2.related_work.tex
\section{Related Work} \label{sec:related-work}

\subsection{Time Series Foundation Models}
Foundation Models (FMs), otherwise known as large pre-trained models, are known to have helped achieve state-of-the-art performance with large language models (LLM’s) in natural language processing (NLP) and advanced models in computer vision (CV)~\cite{shi2024time}. These models consume large amounts of data and serve to provide a “general purpose” representation which may be modified to perform on a diverse set of tasks. As of late, the ideas behind FMs have been extended towards the time series domain allowing for the creation of many different time series foundation models (TSFMs). 

A comprehensive taxonomy provided by Yuxuan Liang’s survey on time series allows us to distinguish and classify TSFMs through the use of four hierarchical levels: data category, model architecture, pre-training techniques, and application domain~\cite{liang2024foundation}. Data category is separated into 3 subsections in which time series are classified as either standard, spatial, or “other” (trajectory/event data). The defining feature of a TSFM, the architecture, is similarly divided into Transformer-based, non-Transformer based, and Diffusion-based models. Most popular and well received models for time series forecasting seem to utilize a transformer based backbone as they are able to handle sequential data efficiently~\cite{miller2024survey}. There exist some models that aim to bridge the gap between different architectures by combining multiple different backbones such as TimeDiT, a novel diffusion transformer-based foundation model for time series analysis~\cite{cao2024timedit}. Lastly, the majority of TSFMs can be categorized as using either self-supervised or fully-supervised pre-training techniques. 

Many papers have proposed models satisfying the above taxonomy, but some notable TSFMs include Lag-Llama and TimeGPT. Both models train on large amounts of diverse data and demonstrate that large-scale pretraining is able to improve forecasting accuracy and adaptability~\cite{rasul2023lagllama}.

\subsection{LLMs for Time Series Forecasting}

Large Language Models (LLMs) have recently begun to demonstrate their prowess in time series analysis and forecasting, stemming from their ability to handle complex numerical sequences across wide domains. A key innovation introduced by LLMs is their ability to bridge the modality gap between textual and numerical data, which allows for time series to be reinterpreted as language modeling tasks.

Early efforts like PatchTST~\cite{nie2022time} demonstrated that breaking a time series down into smaller segments, localized “patch” tokens, significantly helps to enhance the forecasting accuracy while reducing computational overhead. These ideas have been used as a foundation and extended into new models like TST~\cite{zerveas2020transformer}, which work by embedding patches into latent vectors to use as a form of textual representations for numerical inputs.

Many recent methods involve converting real-valued time series into some form of discrete tokens. Chronos~\cite{ansari2024chronos} is an example where the model applies quantization and scaling to transform observations into a fixed vocabulary useful for zero-shot and transfer learning. Similarly, digit-level tokenization~\cite{gruver2024large} is another method in which each digit of a number is treated as a separate token, allowing LLMs to leverage their advantages in next-token prediction abilities for forecasting tasks. Building even further upon this idea, PromptCast~\cite{xue2023promptcastnewpromptbasedlearning} reformulates forecasting as sentence-to-sentence generation using specialized prompt engineering. Models like GPT4TS~\cite{zhou2023one} combine tasks such as anomaly detection, classification, and forecasting through textual prompting alone.

Although there are numerous benefits presented by the usage of LLMs in TSF, they also seem to face challenges when applied to diverse time series data. This is especially noticeable in data with missing values and irregular patterns. To address concerns such as these, LLM4TS~\cite{chang2024llm4tsaligningpretrainedllms} introduces a two-stage training pipeline. This model first works by aligning pretrained LLMs to a standardized time series structure and then begins the fine-tuning for forecasting. Other approaches like TEMPO~\cite{cao2024tempopromptbasedgenerativepretrained} explicitly decompose time series into trend, seasonal, and residual components, enhancing model interpretability and accuracy. 

Furthermore, some architectures have merged different methodologies into a hybrid form in an attempt to draw out more of the LLMs capabilities. TPLLM~\cite{ren2024tpllmtrafficpredictionframework} incorporates CNNs and GCNs alongside LLMs to capture spatial-temporal dependencies in traffic prediction. GenTKG~\cite{liao2024gentkg} leverages retrieval-augmented generation and parameter-efficient tuning for temporal knowledge graphs.

Despite the previous works done in adapting LLMs for time series, many existing and recent methods seem to require heavy fine-tuning and/or a complex architecture that is usually associated with high costs. In this work, we propose PatchInstruct, a simple and straightforward prompting framework that divides time series data into smaller segments called patches, and directly prompts LLMs without the need for any additional training. This approach minimizes token usage while maintaining a strong predictive performance on low horizons across diverse domains.

%% file: draft/3.method.tex
\section{Methodology}

We propose an approach that leverages Large Language Models (LLMs) for time series forecasting through specialized prompt engineering techniques that eliminates the need for model fine-tuning or architectural modifications. 

Our approach begins with a zero-shot baseline, inspired by TimeLLM \cite{gruver2024large}, where the model is prompted with raw historical time series values and tasked with predicting future values. While this baseline offers simplicity and generality, it lacks the inductive bias necessary to capture local temporal dynamics, leading to suboptimal performance in complex forecasting settings. Our second baseline S$^2$IP-LLM introduces significant inference-time overhead due to its reliance on complex decomposition pipelines and fine-tuning, limiting its scalability in real-world deployments.

\begin{figure*}[t]
  \centering
  \includegraphics[scale=0.95]{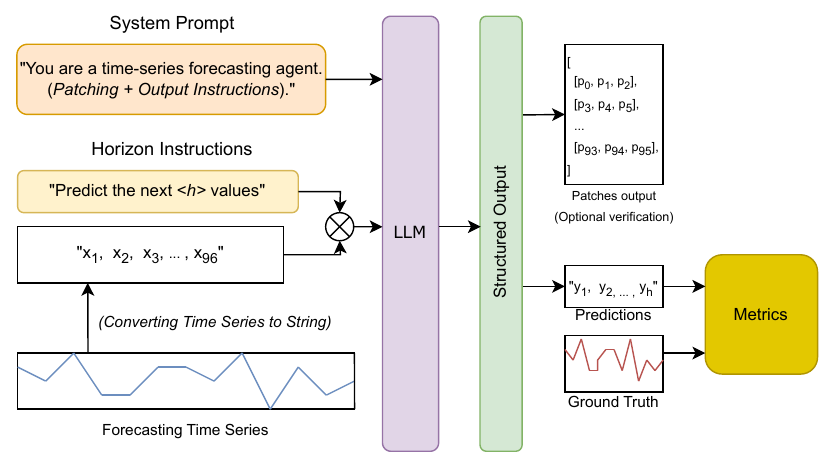}
  \caption{PatchInstruct Forecasting Pipeline}
  \label{fig:eval_figure}
\end{figure*}

To address these limitations, we introduce PatchInstruct (see Figure \ref{fig:eval_figure}), a prompting strategy that encode temporal structure through patch-based representations, and provide pretrained LLMs more context on the dataset. The core idea is to decompose a time series into fixed-length overlapping patches and provide them to the LLM in a structured format, along with instructions to predict future values.

Additionally, we experimented the model's forecasting ability by supplementing the target time series with a small set of similar time series referred to as Neighbors (Neighs). Specifically, we select the five most similar time series from the dataset's past seen data, referred to as neighbors. The motivation behind this approach is to provide the LLM with additional contextual signals and recurring patterns that may not be fully observable in the target series alone.

We finally also tested a combination of these the Patch-Instruct and Neighbors strategy, enriching the prompt with structurally decomposed information from the target series (via patching), while augmenting it with relevant patterns from similar series (via nearest neighbors).

In the following subsections, we detail the construction of each approach, describe the datasets and evaluation metrics used, and present a comparative analysis of their forecasting performance.

We evaluate our method on two time series datasets: Weather and Traffic. These datasets comprised of continuous measurements sampled at regular intervals. A 96-timestep input window is used to forecast future horizons of 1,2,3,4,5,6 and 12 steps.

\input{draft/prompt}

\subsection{Evaluation}
We evaluate forecasting performance using Mean Squared Error (MSE), Mean Absolute Error (MAE). Runtime Efficiency and Input/Output token usage.

%% file: draft/prompt.tex
\subsection{Overview of Framework}

\ryan{Can show algorithm for this too, that takes as input time-series, decomposes it into patches, then gives those patches as input into another component that combines these things. }

Our framework is designed to adapt large language models (LLMs) for time series forecasting without any fine-tuning, using carefully structured prompts that condition the model with temporal data and forecasting instructions. The pipeline is modular and supports multiple prompting strategies, including PatchInstruct, Neighbors, and PatchInstruct + Neighbors, and Zeroshot by modifying the structure of the system prompt and the input representation.

At inference time, a raw time series is converted into a sequence of string-formatted numerical values. Depending on the method, additional transformations are applied—for example, decomposing the sequence into overlapping fixed-length patches (in PatchInstruct), or retrieving similar time series (in Neighs). These inputs are concatenated with forecasting instructions (e.g., "Predict the next h values") and passed to the LLM. The output is parsed into a numerical forecast and compared with the ground truth using standard forecasting metrics.

In addition to forecasting, PatchInstruct also prompts the model to output reconstructed patches from the input, enabling an optional interpretability step. These predicted patches can be compared with the actual ones to assess whether the LLM is learning meaningful temporal structure and capturing local dynamics, thus providing deeper insight into the model’s understanding of the task.

\subsection{Prompt Design}
We now delve deeper into the specific construction of each prompting strategy. This section provides detailed formulations illustrating how time series data, patching instructions, and neighboring trends are encoded within the input to the LLM. The prompts were designed through rigorous empirical testing to ensure clarity and effectiveness. Each prompt consists of a system prompt, which defines the forecasting method and describes how we construct the series, and a user prompt, which contains the actual time series data provided to the LLM for prediction.

PatchInstruct is built upon the zeroshot prompt inspired by LLMTime, this method decomposes the time series into patches and the LLM uses them to form predictions. Below, we outline the structure of the best Patch-Instruct prompt. Additional experiments exploring alternative patching strategies are presented in the Appendix \ref{sec:prompt-results}. The following prompt is a user prompt used for most methods to specify the horizon, and give the context window to the LLM.

\begin{PromptTextBox}{Horizon Prompt (Input Time Series)}
Continue the following sequence without producing any additional text. Sequence: \textit{<$x_1, x_2, x_3,..., x_{96}$>}. Predict the next 3 values.
\end{PromptTextBox}

\begin{PromptTextBox}{PatchInstruct System Prompt}
You are a forecasting assistant that sees time series data. The sequence represents 
the total regional humidity measured every 10 minutes.

Task: (1) Split the series into overlapping patches with window size 3 and stride 1. (2) Generate the patches in natural order, then reverse the list so the most recent patch appears first. (3) Use these patch tokens to forecast the next 3 values.

Output format:

\texttt{Patches:}

\texttt{[}\texttt{[latest\_patch], ..., [oldest\_patch]}\texttt{]}

\texttt{Prediction:}

\texttt{[y1, y2, y3]}

No headings or extra words.  
Decimals \(\leq\) 4 places; keep leading zeros (e.g., \texttt{0.8032}).
\end{PromptTextBox}

For the System prompt, we included both the patching instructions and dataset-specific information, such as the name of the series. Among the variants of prompts tested, we found reverse patching to perform the best. Further details and comparisions of patching strategies are provided in (Section~\ref{sec:prompt-results}; see appendix for Table \ref{tab:ablation_patchInstruct}).

Neighs is built upon our zero-shot prompt. This method adds closest neighboring series in terms of euclidean distance over all the past windows of data and construct a composite prompt by giving all the 5 neighboring prompts as additional context. We outline the structure of the Neighs prompt used for the weather dataset in the ``Neighs System Prompt'' box. We specified the number of neighbors that, and gave the model additional instructions. 

PatchInstruct+Neighs integrates the strengths of both PatchInstruct and Neighs approaches. We combined the two methods using the system prompt in the ``PatchInstruct+Neighs system prompt'' box.

\begin{PromptTextBox}{Neighs System Prompt}
You are a forecasting assistant that sees time series data. The sequence represents 
the total regional humidity measured every 10 minutes. You will also be given 5 neighbor time-series similar to the one to forecast. Use it to understand the trends.

Output format:
\texttt{[y1, y2, y3]}

No headings or extra words.  
Decimals \(\leq\) 4 places; keep leading zeros (e.g., \texttt{0.8032}).

\end{PromptTextBox}

\begin{PromptTextBox}{PatchInstruct + Neighs System Prompt}
You are a forecasting assistant that sees time series data. The sequence represents the total regional humidity measured every 10 minutes. You will also be given 5 neighbor time-series similar to the one to forecast. Use it to understand the trends.

Task: (1) Split the series into overlapping patches with window size 3 and stride 1. (2) Generate the patches in natural order, then reverse the list so the most recent patch appears first. (3) Use these patch tokens to forecast the next 3 values.

Output format:

\texttt{Patches:}

\texttt{[}\texttt{[latest\_patch], ..., [oldest\_patch]}\texttt{]}

\texttt{Prediction:}

\texttt{[y1, y2, y3]}

No headings or extra words.  
Decimals \(\leq\) 4 places; keep leading zeros (e.g., \texttt{0.8032}).
\end{PromptTextBox}

\noindent
In summary, these prompt-based variants allow us to systematically assess the impact of explicit instructions for time series decomposition, patching, and neighbor augmentation within LLM-based forecasting frameworks. 
The results, presented in Table~\ref{tab:patchneighs}, provide a comparative analysis of these prompting strategies.

%% file: draft/4.1.dataset.tex
\section{Experiments}\label{sec:exp}
In this section, we design experiments to investigate our proposed patch instruct framework.

\subsection{Datasets}

We evaluate our approach on two real-world datasets: Weather and Traffic. Weather captures fast-changing environmental conditions, while Traffic reflects urban flow patterns with spatial and temporal dependencies

The {Weather} dataset is collected from a meteorological station at the Max Planck Institute for Biogeochemistry (Jena, Germany). It contains 14 meteorological features, including temperature, humidity, and atmospheric pressure measurements. The high-frequency recordings capture intricate weather dynamics critical for testing short-term forecasting precision.

The {Traffic} dataset consists of sensor network data from Los Angeles, collected between March and June 2012. It records traffic flow rates and congestion patterns across urban arteries. The spatial-temporal correlations in this dataset test the model's ability to capture complex topological dependencies in transportation systems.

We summarize key statistics in Table \ref{tab:datasets_statistics}. This selection provides systematic coverage of (1) different sampling frequencies, (2) variable sequence lengths, and (3) heterogeneous feature interactions---three critical axes for stress-testing tokenization strategies in temporal learning tasks. The datasets' public availability ensures reproducibility, while their domain diversity demonstrates our method's generalizability beyond narrow application contexts.

\begin{table}[ht!]
\centering
\fontsize{8pt}{8pt}\selectfont
\setlength{\tabcolsep}{4pt}
\renewcommand{\arraystretch}{1.25}
\resizebox{1.0\linewidth}{!}{
\begin{tabular}{lccccc}
\toprule
\textbf{Dataset} & \textbf{Features} & \textbf{Frequency} & \textbf{Time Span} & \textbf{Samples} & \textbf{Value Range} \\
\midrule
\textsc{Weather} & 14 & 10 minutes & 3 years & 157,680 & 0.5--18.13 \\
\textsc{Traffic} & 181 & Hourly & 4 months & 34,172 & 2.5--70 \\
\bottomrule
\end{tabular}
}
\caption{Summary of datasets used in our experiments.}
\label{tab:datasets_statistics}
\end{table}

\subsection{Main Results} \label{sec:main-results}

For our experiments, we adopt S$^2$IP-LLM as the primary baseline, a method that aligns time series embeddings with the semantic space of a pre-trained LLM through a tokenization framework. While effective, S$^2$IP-LLM suffers from significant computational overhead, requiring extensive training and inference time due to its fine-tuning of LLM components. All methods are evaluated in a consistent zero-shot setting without model retraining to isolate the impact of prompting strategies.

Using various prompting strategies, we instructed a pre-trained LLM to consider time-series patches and utilize them for forecasting without any additional fine-tuning or retraining. Our experiments demonstrate that such patch-based prompting methods can significantly improve forecasting performance across multiple datasets and over shorter horizons. In contrast to models like S$^2$IP-LLM, which rely on explicit decomposition, semantic alignment, and parameter tuning, our approach leverages instruction-tuned LLMs. Among all the strategies evaluated the PatchInstruct technique consistently delivered the best results. This suggests that prompting pre-trained LLMs with thoughtfully structured temporal context can match or even surpass models trained from scratch, offering a lightweight yet effective alternative for time-series forecasting.

Table \ref{tab:token_time_grouped} presents a performance comparison between S$^2$IPLLM (the baseline) and our best-performing Patch Instruct method across multiple time series datasets and forecast horizons. The results clearly indicate that Patch Instruct consistently outperforms the baseline in terms of both MSE and MAE. Finally, Table \ref{tab:token_time_grouped} compares the two methods in terms of input/output token counts and computation time. The analysis reveals that Patch Instruct not only improves forecasting accuracy but also significantly reduces computational overhead. Overall, this comparison highlights the efficiency and effectiveness of the Patch Instruct method.

\begin{table}[ht!]
\fontsize{8pt}{8pt}\selectfont
\setlength{\tabcolsep}{2pt}
\renewcommand{\arraystretch}{1.25}
\resizebox{1.0\linewidth}{!}{
\begin{tabular}{c@{}c cc cc cc}
\toprule
\textbf{Dataset} & \textbf{Horizon} & 
\multicolumn{2}{c}{\textbf{S$^{2}$IP-LLM}} &
\multicolumn{2}{c}{\textbf{Zeroshot}} & 
\multicolumn{2}{c}{\textbf{PatchInstruct}} \\
\cmidrule(l){3-4}
\cmidrule(l){5-6}
\cmidrule(l){7-8}
& & \textbf{MSE} & \textbf{MAE} & \textbf{MSE} & \textbf{MAE} & \textbf{MSE} & \textbf{MAE} \\
\midrule
\multirow{7}{*}{\sc Weather} 
& 1 & 0.0095 & 0.056 & 0.0028 & 0.043 & \textbf{0.0014} & \textbf{0.029} \\
& 2 & 0.017 & 0.077 & 0.0085 & 0.072 & \textbf{0.0076} & \textbf{0.067} \\
& 3 & 0.0238 & 0.0875 & \textbf{0.0106} & \textbf{0.068} & 0.0110 & 0.085 \\
& 4 & 0.0326 & 0.1051 & \textbf{0.0115} & \textbf{0.085} & 0.0236 & 0.113 \\
& 5 &  0.0371 & 0.1120 & 0.0277 & 0.1 & \textbf{0.0159} & \textbf{0.094} \\
& 6 & 0.0439 & 0.1228 & 0.0204 & 0.11 & \textbf{0.0101} & \textbf{0.083} \\
& 12 & 0.0904 & 0.1823 & 0.1098 & 0.221 & \textbf{0.0436} & \textbf{0.137} \\
\midrule
\multirow{7}{*}{\sc Traffic}
& 1 & 21.0814 & \textbf{2.4067} & 43.49 & 3.18 & \textbf{20.05} & 2.76 \\
& 2 & 24.0935 & 2.4919 & 23.47 & 2.53 & \textbf{9.38} & \textbf{1.89} \\
& 3 & 29.9573 & 2.7849 & 22.38 & 2.54 & \textbf{6.47} & \textbf{1.78} \\
& 4 & 29.8382 & 2.6147 & 27.50 & 2.87 & \textbf{11.15} & \textbf{1.89} \\
& 5 & 36.0289 & 2.7971 & 34.27 & 3.20 & \textbf{8.46} & \textbf{1.88} \\
& 6 & 42.3193 & 3.0444 & 29.66 & 3.07 & \textbf{25.59} & \textbf{2.75} \\
& 12 & \textbf{68.7149} & \textbf{3.8473} & 296.20 & 7.72 & 235.75 & 5.89 \\
\bottomrule
\end{tabular}}
\caption{Results comparing our approach to baselines.}
\label{tab:univariate_results}
\end{table}

\begin{table}[ht!]
\fontsize{8pt}{8pt}\selectfont
\setlength{\tabcolsep}{2pt}
\renewcommand{\arraystretch}{1.25}
\resizebox{1.0\linewidth}{!}{
\begin{tabular}{c@{}c ccc ccc ccc}
\toprule
\textbf{Dataset} & \textbf{Horizon} & 
\multicolumn{3}{c}{\textbf{S$^{2}$IP-LLM}} & 
\multicolumn{3}{c}{\textbf{Zeroshot}} & 
\multicolumn{3}{c}{\textbf{PatchInstruct}} \\
\cmidrule(l){3-5}
\cmidrule(l){6-8}
\cmidrule(l){9-11}
& & \textbf{Time (s)} & \textbf{IT} & \textbf{OT} & \textbf{Time (s)} & \textbf{IT} & \textbf{OT} & \textbf{Time (s)} & \textbf{IT} & \textbf{OT} \\
\midrule
\multirow{7}{*}{\sc Weather}
& 1 & 535.42 & 7 & 1 & 1.36 & 7370 & 80 & 1.24 & 8500 & 80 \\
& 2 & 518.45 & 7 & 2 & 1.06 & 7370 & 120 & 1.20 & 8500 & 120 \\
& 3 & 533.73 & 7 & 3 & 1.20 & 7370 & 160 & 1.01 & 8500 & 160 \\
& 4 & 518.37 & 7 & 4 & 1.01 & 7370 & 200 & 1.16 & 8500 & 196 \\
& 5 & 537.85 & 7 & 5 & 1.14 & 7370 & 240 & 1.29 & 8500 & 216 \\
& 6 & 522.43 & 7 & 6 & 1.35 & 7370 & 280 & 2.05 & 8500 & 280 \\
& 12 & 558.67 & 7 & 12 & 1.44 & 7370 & 520 & 1.51 & 8500 & 499 \\
\midrule
\multirow{7}{*}{\sc Traffic}
& 1 & 50.94 & 7 & 1 & 2.59 & 7360 & 80 & 1.31 & 7950 & 86 \\
& 2 & 52.88 & 7 & 2 & 2.04 & 7360 & 116 & 1.05 & 7950 & 134 \\
& 3 & 55.34 & 7 & 3 & 1.17 & 7360 & 160 & 1.11 & 7950 & 185 \\
& 4 & 51.00 & 7 & 4 & 2.13 & 7360 & 200 & 1.17 & 7950 & 223 \\
& 5 & 49.96 & 7 & 5 & 1.14 & 7360 & 240 & 1.12 & 7950 & 239 \\
& 6 & 50.11 & 7 & 6 & 1.38 & 7360 & 268 & 1.23 & 7950 & 336 \\
& 12 & 52.66 & 7 & 12 & 1.78 & 7360 & 520 & 1.36 & 7950 & 558 \\
\bottomrule
\end{tabular}}
\caption{Token and Time Comparison for Forecasting.}
\label{tab:token_time_grouped}
\end{table}

\subsection{Cost vs. Performance Analysis} \label{sec:results-tradeoffs}

Our instruction-based forecasts deliberately spend more input tokens than the baseline S$^2$IP-LLM. Across the prompt variants, a single prediction consumes about \(\approx\) 800 - 1000 input tokens. By contrast, S$^2$IP-LLM needs only the horizon-length of output tokens once its patch encoder has been trained. The extra prompt length therefore represents about a 100 times increase in  in front-loaded cost.

Because our method relies on an already-trained LLM and does no task-specific fine-tuning, the end-to-end latency of producing a forecast collapses from minutes to just seconds.  
For example, on the Weather dataset at horizon \(=1\), S$^2$IP-LLM requires 535 s, whereas Reverse Patch returns the prediction in \(0.86\) (see Table ~\ref{tab:token_time_grouped}).  
Thus, even after accounting for the larger prompt, our approach is two to three orders of magnitude faster in real-time settings.

The additional 800--900 input tokens result in a substantial improvement in short-range forecasting accuracy.  
On Weather (H=1), mean squared error (MSE) drops from \(1.15\times10^{-2}\) to \(2.6\times10^{-4}\) (a 97.7\% reduction), and mean absolute error (MAE) decreases from \(6.52\times10^{-2}\) to \(1.4\times10^{-2}\).  
Similar improvements are observed on Traffic, where the MSE at the same horizon is reduced by 85\%, indicating that the gains generalize across domains.

Given (i) the low marginal price of LLM tokens relative to GPU training hours, and (ii) the consistent short-horizon error reductions that are operationally most valuable, the accuracy and latency benefits comfortably offset the larger prompt size. Hence trading cheap tokens for immediate, higher-quality forecasts yields a more favorable cost–performance envelope than the current state of the art, especially when rapid deployment and low engineering overhead are priorities.

\subsection{Neighbor Results} \label{sec:patch-instruct-neigh}
In order to understand whether incorporating neighboring time-series into our PatchInstruct approach leads to better performance we compare our results for PatchInstruct and Neighs across multiple datasets (Table~\ref{tab:patchneighs}).

Both the Neighs and PatchInstruct+Neighs prompting strategies demonstrate clear improvements over the S$^2$IP-LLM baseline across datasets. As seen in Table \ref{tab:patchneighs}, using Neighs alone often improves performance over PatchInstruct, particularly in the Weather dataset. For example, at horizon 2, Neighs reduces the MSE from 0.0076 (PatchInstruct) to 0.0039 and MAE from 0.067 to 0.051. At horizon 4, Neighs again performs better with an MSE of 0.0172 compared to 0.0236. These gains suggest that incorporating neighboring series can help the model infer more accurate trends by providing contextual information beyond the target sequence itself.

However, this is not universally true. In some cases, Neighs and PatchInstruct+Neighs underperform compared to PatchInstruct. For instance, in the Traffic dataset at horizon 5, Neighs shows a significant degradation, increasing the MSE from 8.46 (PatchInstruct) to 43.50, and PatchInstruct+Neighs to 36.43. This indicates that when neighbor series are less correlated, they can introduce confusion rather than useful context.

Despite these exceptions, the PatchInstruct+Neighs strategy still achieves the best overall performance in many cases aswell. But these results also highlight the importance of carefully selecting relevant neighbors to avoid negative transfer and ensure consistent forecasting improvements.

These results underline the strength of combining temporal structuring (via patches) with spatial context (via neighbors), enabling the model to learn more holistic representations and deliver significantly more accurate forecasts than our baseline.

\begin{table}[ht!]
\fontsize{8pt}{8pt}\selectfont
\setlength{\tabcolsep}{2pt}
\renewcommand{\arraystretch}{1.25}
\resizebox{1.0\linewidth}{!}{
\begin{tabular}{c@{}c cc cc cc}
\toprule
\textbf{Dataset} & \textbf{Horizon} & 
\multicolumn{2}{c}{\textbf{PatchInstruct}} & 
\multicolumn{2}{c}{\textbf{Neighs}} & 
\multicolumn{2}{c}{\textbf{PatchInstruct+Neighs}} \\
\cmidrule(l){3-4}
\cmidrule(l){5-6}
\cmidrule(l){7-8}
& & \textbf{MSE} & \textbf{MAE} & \textbf{MSE} & \textbf{MAE} & \textbf{MSE} & \textbf{MAE} \\
\midrule
\multirow{7}{*}{\sc Weather}
& 1  & \textbf{0.0014} & \textbf{0.029} & 0.0024 & 0.042 & 0.0032 & 0.046 \\
& 2  & 0.0076 & 0.067 & \textbf{0.0039} & \textbf{0.051} & 0.0056 & 0.056 \\
& 3  & 0.0110 & 0.085 & \textbf{0.0083} & \textbf{0.065} & 0.0138 & 0.087 \\
& 4  & 0.0236 & 0.113 & 0.0172 & 0.091 & \textbf{0.0114} & \textbf{0.075} \\
& 5  & 0.0159 & 0.094 & \textbf{0.0105} & \textbf{0.077} & 0.0124 & 0.088 \\
& 6  & \textbf{0.0101} & \textbf{0.083} & 0.0116 & 0.084 & 0.0338 & 0.108 \\
& 12 & 0.0436 & \textbf{0.137} & \textbf{0.0371} & 0.144 & 0.0393 & 0.141 \\
\midrule
\multirow{7}{*}{\sc Traffic}
& 1  & \textbf{20.05}  & 2.76  & 35.15  & 3.05  & 22.09  & \textbf{2.72} \\
& 2  & \textbf{9.38}   & \textbf{1.89}  & 18.85  & 2.50  & 15.40  & 2.19 \\
& 3  & \textbf{6.47}   & \textbf{1.78}  & 26.67  & 2.74  & 13.61  & 2.10 \\
& 4  & \textbf{11.15}  & \textbf{1.89}  & 21.41  & 2.57  & 13.22  & 2.15 \\
& 5  & \textbf{8.46}   & \textbf{1.88}  & 43.50  & 3.39  & 36.43  & 3.25 \\
& 6  & 25.59  & 2.75  & 40.42  & 3.35  & \textbf{14.94}  & \textbf{2.13} \\
& 12 & \textbf{235.75} & \textbf{5.89}  & 285.82 & 7.71  & 269.68 & 6.88 \\
\bottomrule
\end{tabular}}
\caption{Forecasting Comparison: PatchInstruct vs Neighs vs PatchInstruct+Neighs.}
\label{tab:patchneighs}
\end{table}

%% file: draft/4.2.exp.tex
\section{Analysis}

This analysis compares the performance of S$^2$IP-LLM (baseline) against our approach across different datasets and forecast horizons. The comparison focuses on error metrics (MSE and MAE) where lower values indicate better performance.

\subsection{Performance Overview Across Models}

The S$^2$IP-LLM baseline consistently shows higher error rates compared to our methods across datasets. Our methods shows remarkable improvements, with percentage reductions in MSE ranging from approximately 13\% to 85\% depending on the dataset and method. For the Weather dataset, all our methods achieve over 80\% MSE improvement. Our method achieve orders-of-magnitude faster runtimes than S$^2$IP-LLM with modest token usage growth, making them highly efficient for inference, especially when balanced with patch or neighbor-based prompts.
\\
\subsection{Method-Specific Performance}
Our approach exhibits distinct strengths across datasets and forecasting horizons. The PatchInstruct framework demonstrates the most balanced performance, delivering substantial improvements on both the Traffic and Weather datasets—achieving up to 83\% and 85\% improvement over the baseline, respectively. The Neighs variant, which augments prompts with the closest neighboring time series, performs particularly well on the Weather dataset. Meanwhile, the combined PatchInstruct+Neighs strategy outperforms other methods on the Traffic dataset at longer horizons, highlighting the benefit of incorporating both local structure and external context in more challenging settings. These results suggest that method selection can be guided by the characteristics of the dataset and the specific forecasting task, with PatchInstruct offering a robust default across most conditions.

\subsection{Dataset-Specific Analysis}

For Weather forecasting, all methods substantially outperform the baseline. Overall MSE values are reduced from 0.009--0.0904 (baseline) to as low as 0.0014--0.043 (our methods).

 PatchInstruct deliver substantial improvements, reducing MSE from 21--68 (baseline) to 6.47--20.05. However, Neighs and PatchInstruct+Neighs perform poorly in this scenario.

Across both approaches, forecast accuracy generally decreases as the horizon increases, but this pattern varies by dataset and method. For the Weather dataset, the performance degradation with longer horizons is less pronounced, especially for PatchInstruct+Neighs in multivariate settings.

\subsection{Key Insights and Implications}

The optimal forecasting method varies notably depending on the characteristics of the dataset and the forecasting horizon. For the Weather dataset, the PatchInstruct and Neighs strategy yields the most accurate results, effectively capturing the contextual signals from related series. In contrast, for the Traffic dataset, our main approach PatchInstruct performs best, suggesting that more complex augmentation may not always be beneficial in settings with high variability or less correlated neighbors.

\section{Conclusion}

The analysis demonstrates that prompt-based methods generally outperform the S$^2$IP-LLM baseline across most forecasting scenarios, especially at shorter horizons. The optimal method depends significantly on the specific dataset and forecast horizon, with PatchInstruct dominating in the majority of cases. This suggests that while prompt-based strategies offer a lightweight and effective alternative for time series forecasting.

\section{Limitations}
While our method achieves competitive accuracy compared to the S$^2$IP-LLM baseline, several limitations warrant consideration. First, the evaluation is limited to two benchmark datasets, which, though diverse, may not fully represent the diversity of real-world time series scenarios, such as irregular sampling or high-frequency patterns. Second, the framework remains heavily contingent upon carefully engineered prompts, introducing a labor-intensive design process that risks overfitting to specific tasks or datasets without systematic adaptation strategies. Future research should prioritize expanding dataset coverage, and developing adaptive prompting mechanisms.

%% file: draft/appendix.tex
\appendix
\section{Summary of Forecasting Results Across different datasets}\label{sec:prompt-results}

\begin{table*}[ht!]
\fontsize{8pt}{8pt}\selectfont
\setlength{\tabcolsep}{2pt}
\renewcommand{\arraystretch}{1.25}
\caption{Ablation Study: Comparing Variants of Patch-Based Prompting Strategies Across Datasets.}
\label{tab:ablation_patchInstruct}
\resizebox{1.0\linewidth}{!}{
\begin{tabular}{c@{}c ccc ccc ccc ccc ccc}
\toprule
\textbf{Dataset} & \textbf{Horizon} & 
\multicolumn{3}{c}{\textbf{Basic}} & 
\multicolumn{3}{c}{\textbf{Non-Overlapping}} & 
\multicolumn{3}{c}{\textbf{STR Decompose}} & 
\multicolumn{3}{c}{\textbf{Reverse Patches}} & 
\multicolumn{3}{c}{\textbf{Meta Patches}} \\
\cmidrule(l){3-5}
\cmidrule(l){6-8}
\cmidrule(l){9-11}
\cmidrule(l){12-14}
\cmidrule(l){15-17}
& & \textbf{MAE} & \textbf{MSE} & \textbf{Time} & \textbf{MAE} & \textbf{MSE} & \textbf{Time} & \textbf{MAE} & \textbf{MSE} & \textbf{Time} & \textbf{MAE} & \textbf{MSE} & \textbf{Time} & \textbf{MAE} & \textbf{MSE} & \textbf{Time} \\
\midrule
\multirow{3}{*}{\sc Weather}
& 1 & 0.014 & 0.0003 & 0.66 & 0.012 & 0.0002 & 1.6151 & \textbf{0.009} & \textbf{0.0001} & 1.2553 & 0.015 & 0.0005 & 1.2290 & 0.020 & 0.0007 & 0.7471 \\
& 3 & 0.050 & 0.0045 & 0.989 & 0.055 & 0.0064 & 1.3580 & \textbf{0.045} & \textbf{0.0030} & 1.0737 & 0.053 & 0.0045 & 0.9732 & 0.067 & 0.0073 & 0.9269 \\
& 6 & 0.078 & 0.0116 & 1.146 & 0.063 & 0.0079 & 3.9016 & 0.100 & 0.0230 & 1.1523 & \textbf{0.056} & \textbf{0.0070} & 1.5120 & 0.060 & 0.0074 & 0.9760 \\
\midrule
\multirow{3}{*}{\sc Traffic}
& 1 & 1.43 & 6.27 & 0.699 & 1.35 & 5.60 & 1.3404 & 1.34 & 4.17 & 1.2677 & \textbf{1.15} & \textbf{3.69} & 1.1221 & 1.32 & 5.35 & 1.3846 \\
& 3 & 1.07 & 3.36 & 0.940 & 1.47 & 7.17 & 1.1910 & 0.99 & 2.53 & 1.1801 & \textbf{0.89} & \textbf{1.83} & 1.2018 & 0.94 & 2.38 & 1.1584 \\
& 6 & 1.14 & 4.37 & 0.910 & 1.14 & 3.60 & 1.3010 & 1.79 & 8.71 & 1.5667 & \textbf{0.89} & \textbf{2.26} & 1.1137 & 1.03 & 2.95 & 1.1650 \\
\bottomrule
\end{tabular}
}
\end{table*}

Table \ref{tab:ablation_patchInstruct} represents evaluating five prompting strategies—Basic, Non-Overlapping, STR Decompose, Reverse Patches, and Meta Patches—across the Weather and Traffic datasets, for horizons of 1, 3, and 6. While STR Decompose occasionally shows the lowest error for short-term predictions, Reverse Patch Instruct consistently delivers strong performance across all horizons and datasets. Notably, in long-range forecasts (H=6), where prediction becomes more challenging, Reverse Patch Instruct achieves the lowest or near-lowest MAE and MSE in both datasets, highlighting its stability and generalizability. Although it incurs a slightly higher inference time than the most lightweight methods, the trade-off is minimal when weighed against the accuracy benefits. Overall, the results suggest that Reverse Patch Instruct is the most effective and reliable strategy, outperforming other variants in terms of both robustness and predictive accuracy.

\subsection{Basic PatchInstruct}
The Basic PatchInstruct method employs overlapping sliding windows (size=3, stride=1) to capture local temporal patterns, followed by strategic sequence reversal to prioritize recent context. Unlike conventional approaches that process time series chronologically, this method reverses the generated patches such that the most recent window \texttt{[x\textsubscript{t-2}, x\textsubscript{t-1}, x\textsubscript{t}]} appears first in the token sequence. This architectural innovation forces the model to attend to immediate temporal patterns before historical context, combining the local sensitivity of patch-based methods \cite{nie2022time} with explicit recency prioritization. The approach demonstrates particular efficacy in high-frequency electricity demand forecasting where near-term consumption patterns strongly influence subsequent values.

\subsection{Non-Overlapping PatchInstruct}
This variant utilizes non-overlapping windows where both window size and stride equal the prediction horizon (typically 3). The method partitions the series into discrete blocks like \texttt{[8.35, 8.36, 8.32]} followed by \texttt{[8.45, 8.35, 8.25]}, eliminating redundant data coverage while maintaining temporal progression. The design trades off some contextual granularity for computational efficiency, making it suitable for scenarios with pronounced periodic patterns. By processing patches in natural order without sequence reversal, the method preserves strict temporal causality, particularly effective when historical seasonal trends dominate the forecasting signal.

\subsection{STR Decompose PatchInstruct}
Integrating seasonal-trend-residual decomposition, this method first separates raw values into trend (trend\_t) and residual (residual\_t = series\_t - trend\_t) components. Each time step becomes a composite token \texttt{[T\textsubscript{t}, R\textsubscript{t}]}, enabling joint modeling of long-term trajectories and short-term fluctuations. These dual-aspect tokens are organized into overlapping windows: 
\begin{equation*}
    \texttt{[[T\textsubscript{1},R\textsubscript{1}], [T\textsubscript{2},R\textsubscript{2}], [T\textsubscript{3},R\textsubscript{3}]]}
\end{equation*}
preserving both local context and decomposition characteristics. The architecture explicitly captures multi-scale temporal dynamics, particularly beneficial for electricity demand series containing both gradual load changes and sudden consumption spikes.

\subsection{Reverse Ordered Patches}
Building on basic patch inversion, this method systematically prioritizes recent context through full sequence reversal of overlapping windows. The architectural innovation forces models to process the final patch \texttt{[x\textsubscript{94}, x\textsubscript{95}, x\textsubscript{96}]} first, implementing a "recency-first" attention mechanism. This structural bias proves particularly effective for 10-minute interval forecasting where immediate consumption patterns (last 30 minutes) contain stronger signals than older data. The approach maintains patch-based efficiency while adding temporal prioritization through simple sequence manipulation.

\subsection{Meta Tokens Patches}
This advanced variant enriches temporal representation through explicit time slot encoding. Each value $v_t$ pairs with its absolute position in the daily cycle (0-143 slots) as \texttt{($v_t$; slot\textsubscript{id})}, creating hybrid tokens like \texttt{(8.35;63)}. These meta-tokens are windowed into overlapping patches:
\texttt{[(v1;slot1), (v2;slot2), (v3;slot3)]  \newline
  [(v2;slot2), (v3;slot3), (v4;slot4)] \newline  
  …  \newline
  [(v94;slot94),(v95;slot95),(v96;slot96)]}
enabling joint learning of consumption patterns and their absolute temporal positions. The fixed slot indices provide crucial circadian context, helping disambiguate similar patterns occurring at different times (e.g., morning vs. evening peaks). This method adapts positional encoding strategies from language models to time series, grounding predictions in both value sequences and absolute time references.

\newpage
\begin{PromptTextBox}{Basic PatchInstruct}
You are a forecasting assistant that sees time series data. The sequence represents the total regional humidity measured every 10 minutes.
\newline
Task: \newline (1) Split the series into overlapping patches with window size 3 and stride 1. \newline (2) Generate the patches in natural order, then reverse the list so the most recent patch appears first. \newline (3) Use these patch tokens to forecast the next 3 values.
\newline
\newline
Output format:

\texttt{Patches:}

\texttt{[}\texttt{[latest\_patch], ..., [oldest\_patch]}\texttt{]}
\newline
\newline
\texttt{Prediction:}

\texttt{[y1, y2, y3]}

No headings or extra words.  
Decimals \(\leq\) 4 places; keep leading zeros (e.g., \texttt{0.8032}).
\end{PromptTextBox}

\begin{PromptTextBox}{Non-Overlapping PatchInstruct}
Tokenize the given time-series data into non-overlapping patches where a patch is a contiguous subsequence of the time-series. Ensure to use a fixed window size equal to the Horizon size (e.g., 3) and the stride is equal to the window size. This means that each patch starts exactly where the previous one ends and there will be no overlap. 
Output patches as a list, in order, using square brackets. 
Each patch becomes a token used to represent local temporal patterns. Use the sequence of patches to predict the next value(s).
Below are a few shot examples of non-overlapping patching:
Time series data:
8.35, 8.36, 8.32, 8.45, 8.35, 8.25, 8.20, 8.09, 8.13, 8.00, 7.94, 7.86 \\

Patches generated based on Horizon (3), stride = 3: \newline
[8.35, 8.36, 8.32] \newline
[8.45, 8.35, 8.25] \newline
[8.20, 8.09, 8.13] \newline
[8.00, 7.94, 7.86] \newline

Prediction:
[7.89, 7.97, 7.94]

\end{PromptTextBox}

\begin{PromptTextBox}{STR Decompose PatchInstruct}
You are a forecasting assistant that receives STL-decomposed tokens.

Input: \newline
  - "series": 96 raw numbers (Humidity demand)  
  - "horizon" : 3   (fixed)

Task: \newline
  1. Decompose the series into \newline
  $trend_t$ and $residual_t = series_t - trend_t$
\newline
  2. For each time-step create a pair token:  ($trend_t$ , $residual_t$).  
  \newline
  3. Split the 96 composite tokens into overlapping patches (window = 3, stride = 1).  \newline
  4. Use those patches to forecast the next 3 raw values.
\newline

Output exactly

\texttt{[[T1,R1], [T2,R2], [T3,R3]]  \newline
  [[T2,R2], [T3,R3], [T4,R4]] \newline  
  …  \newline
  [[T94,R94], [T95,R95], [T96,R96]] }
\newline
\newline
\texttt{Prediction:}

\texttt{[y1, y2, y3]}
\newline
No headings or extra words.  
Decimals \(\leq\) 4 places; keep leading zeros (e.g., \texttt{0.8032}).
\end{PromptTextBox}

\begin{PromptTextBox}{Reverse Ordered Patches PatchInstruct}
You are a forecasting assistant that sees time series data. The sequence represents the total regional humidity measured every 10 minutes.
\newline
Input: \newline
  - "series": 96 raw numbers (Humidity, 10-min cadence)   
  - "horizon" : 3 (fixed)
\newline
Task: \newline
  1. Split the series into overlapping patches (window = 3, stride = 1).
\newline
  2. Generate them in natural order, then reverse the list so the most recent patch appears first.
  \newline
  3. Use those patch tokens to forecast the next 3 normalised values. \newline

Output format:

\texttt{Patches:}

\texttt{[}\texttt{[latest\_patch], \newline ... , \newline[oldest\_patch]}\texttt{]}
\newline
\newline
\texttt{Prediction:}

\texttt{[y1, y2, y3]}

No headings or extra words.  
Decimals \(\leq\) 4 places; keep leading zeros (e.g., \texttt{0.8032}).
\end{PromptTextBox}

\begin{PromptTextBox}{Meta tokens Patches PatchInstruct}
You are a forecasting assistant that sees time series data, where each datapoint is paired with its 10-minute slot index within the day. The sequence represents the total regional humidity measured every 10 minutes.

Input: \newline
  - "series": 96 raw numbers (Humidity, 10-min cadence)   
  - "horizon" : 3 (fixed)

Time-slot index
  - A day is divided into 144 slots (0 → 143).
  - slot = floor((60*HH + MM)/10).
    Example: 10:30 → 63   (because 10*60 + 30 = 630; 630/10 = 63).

Token format
  (value ; $slot_{id}$) \newline     
  $slot_{id}$ corresponds to the measurement's clock time
  
Task: \newline
  1. Convert the 96-point series into 96 two-element tokens as above.
\newline
  2. Split the token stream into overlapping patches (window = 3, stride = 1).
  \newline
  3. Use those patches to forecast the next 3 raw demand values. \newline

Output format:

\texttt{[(v1;slot1), (v2;slot2), (v3;slot3)]  \newline
  [(v2;slot2), (v3;slot3), (v4;slot4)] \newline  
  …  \newline
  [(v94;slot94), (v95;slot95), (v96;slot96)] }
\newline
\newline
\texttt{Prediction:}

\texttt{[y1, y2, y3]}

No headings or extra words.  
Decimals \(\leq\) 4 places; keep leading zeros (e.g., \texttt{0.8032}).
\end{PromptTextBox}